\documentclass[10pt,letterpaper]{article}
\PassOptionsToPackage{numbers,sort&compress}{natbib}
\usepackage[preprint]{neurips_2026}
\ifdefined\pdfminorversion\pdfminorversion=7\fi
\usepackage[T1]{fontenc}
\usepackage{booktabs}
\usepackage{tabularx}
\usepackage{booktabs,longtable,array}
\usepackage[utf8]{inputenc}
\usepackage{graphicx}
\usepackage{amsmath,amssymb}
\usepackage{url}
\usepackage{titletoc}
\usepackage{chngcntr}
\usepackage[hidelinks]{hyperref}
\setcitestyle{numbers,square,comma}

\title{EgoSwarm: Learning Decentralized Robot Swarm Intelligence from Human Swarm Demonstrations}
\author{
Ziqiang Liu$^{1,\dagger}$ \quad
Xitong Chen$^{1,\dagger}$ \quad
Min Li$^{1,*}$ \quad
Yiqun Li$^{1}$ \quad
Zhuoping Yin$^{1}$\\
$^{1}$State Key Laboratory of Intelligent Manufacturing Equipment and Technology\\
Huazhong University of Science and Technology\\
$^{\dagger}$Equal contribution \quad
$^{*}$Corresponding author
}

\begin{document}
\ifdefined\XeTeXversion\fi
\maketitle

\begin{abstract}

Advancing from individual embodied intelligence to swarm embodied intelligence is a key step toward scalable, general-purpose embodied intelligence. Complex tasks often require multiple agents to divide responsibilities, coordinate handovers, and manipulate shared objects collaboratively. However, existing robot learning methods rely primarily on individual demonstrations, which provide limited supervision for modeling dependencies among agent actions, while collecting multi-robot demonstrations at scale remains difficult. Human egocentric demonstrations offer a scalable alternative, yet distributed observations from multiple participants cannot directly supervise decentralized policies. We introduce EgoSwarm, the first robot learning framework for learning decentralized multi-robot collaboration from human swarm demonstrations. EgoSwarm collects more than 50 hours of collaborative experience using a robot-free multi-person egocentric system. We introduce the Swarm Interaction Packet (SIP) and Swarm Interaction Token (SIT) to recover cross-agent interaction relations from distributed observations and transform this experience into local supervision consistent with deployment-time information constraints. We further propose SwarmVLA, which transfers knowledge of swarm behavior to decentralized policies conditioned on local observations, local history, and limited neighbor messages. SwarmVLA achieves a mean success rate of 72.2\% across three categories of multi-robot collaborative tasks. Project page: \url{https://xitong-c.github.io/EgoSwarm_Page/}.

\end{abstract}

\begin{figure}[htbp]
\centering
\includegraphics[width=\textwidth]{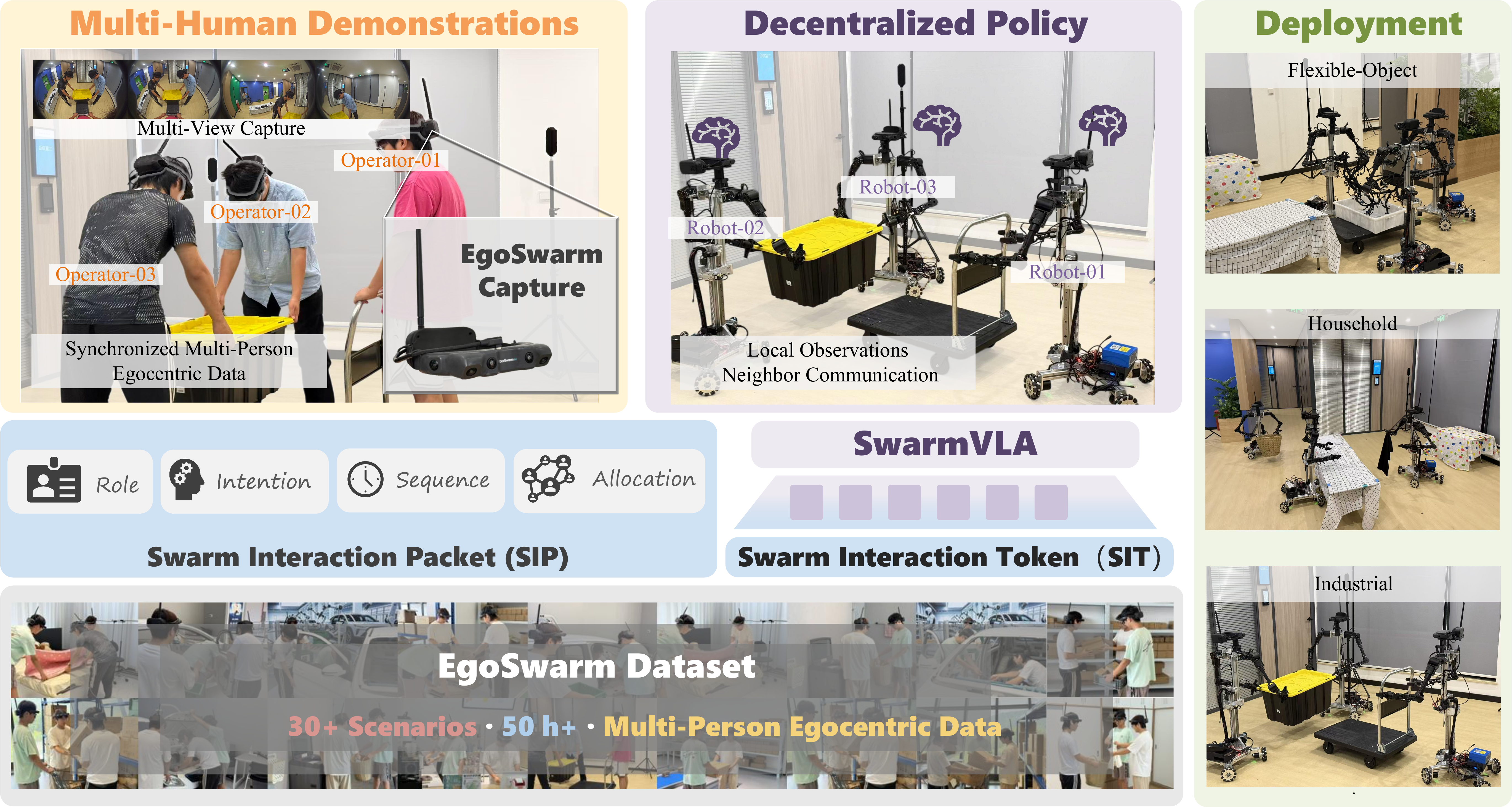}
\caption{\textbf{Overview of EgoSwarm.} EgoSwarm is the first robot learning framework for decentralized multi-robot collaboration from human swarm demonstrations. It reconstructs cross-agent interaction in the form of the Swarm Interaction Packet (SIP), encodes deployment-available interaction information into the Swarm Interaction Token (SIT), and learns locally executable collaboration policies through SwarmVLA.}
\label{fig:1}
\end{figure}

\section{Introduction}\label{sec:intro}

Robotics has long pursued general-purpose embodied intelligence to autonomously perform complex tasks. Many real-world tasks, however, inherently require collaboration among multiple robots. Collaborative transport, multi-participant handovers, and collaborative assembly require robots to divide responsibilities around shared objects and adapt behavior to partner states and task progress. Although vision-language-action (VLA) models have substantially improved generalization in individual manipulation \cite{ref1,ref2}, existing robot learning methods still focus primarily on individual demonstrations and provide limited supervision for inter-agent action dependencies. Multi-robot learning therefore requires more than parallel individual policies: robots must learn coordinated behavior from interaction experience and make decentralized decisions based on locally available information \cite{ref3,ref4,ref5}.

Learning coordinated behavior requires interaction experience at scale. Robot demonstration datasets provide state--action supervision for learning manipulation policies \cite{ref6,ref7}. Scaling teleoperation to multiple robots, however, requires coordinated control across multiple platforms and dedicated supporting infrastructure, making data collection dependent on robot availability and scene-specific setup effort. Humans naturally perform complex collaborative tasks, and egocentric recording offers a more scalable way to capture such real-world experience. Recent large-scale egocentric datasets \cite{ref8,ref9,ref10} and human-to-robot transfer methods \cite{ref11,ref12,ref13,ref14,ref17} have demonstrated that human experience can effectively support robot skill learning. However, existing work primarily focuses on individual behavior. A general learning paradigm for extracting cross-agent interaction relations from multi-person egocentric experience and using them to supervise decentralized multi-robot policies remains underexplored.

Multi-person egocentric experience cannot directly provide supervision for decentralized swarm policies. Individual demonstrations establish a clear correspondence between observations and actions, whereas the information required for collaborative decision-making is often distributed across the local views of different participants. An agent action may depend on partner states, shared objects, and interaction progress, and the relations underlying waiting, handovers, and joint manipulation cannot be recovered by simply combining independent trajectories. Moreover, offline fusion of observations from multiple participants can reconstruct a global view of the swarm state, whereas each deployed robot has access only to local observations, local history, and limited neighbor messages. This creates an information mismatch between training and deployment. Collecting more collaborative data alone does not resolve this mismatch. The central challenge is therefore to recover cross-agent interaction relations from distributed human experience and transform them into local supervision consistent with deployment-time information constraints.

To address this challenge, we propose EgoSwarm, the first robot learning framework for learning decentralized multi-robot collaboration from human swarm demonstrations. EgoSwarm synchronously records natural human collaboration from multiple egocentric views, enabling collaborative experience to be collected at scale without requiring robots during data collection. Using this system, we collect more than 50 hours of human collaborative experience. From these distributed observations, EgoSwarm integrates temporal information, relative geometry, shared-object information, and contact relations into Swarm Interaction Packets (SIPs) with associated quality annotations. It then encodes the interaction information available at deployment from local observations and neighbor messages into Swarm Interaction Tokens (SITs). Building on these representations, we introduce SwarmVLA, which leverages the rich interaction information contained in human swarm demonstrations to learn coordinated behavior and transfer this knowledge to decentralized local policies. During execution, each robot collaborates using only local observations, local history, and limited neighbor messages. Across three categories of real-world multi-robot tasks---household manipulation, industrial manipulation, and flexible wire-harness manipulation---SwarmVLA achieves a mean success rate of 72.2\%, substantially outperforming the evaluated baselines.

Our main contributions are as follows:

\begin{itemize}
\item \textbf{EgoSwarm Capture:} A robot-free multi-person egocentric data collection system that synchronously records human swarm interactions and individual actions.
\item \textbf{SIP and SIT:} Structured interaction representations that convert distributed multi-person observations into supervision consistent with deployment conditions for decentralized multi-robot learning.
\item \textbf{SwarmVLA:} A decentralized policy learning framework that jointly uses human swarm experience and robot demonstrations to learn cross-agent coordination from local observations and limited neighbor communication.
\end{itemize}

\begin{figure}[tbp]
\centering
\includegraphics[width=\textwidth]{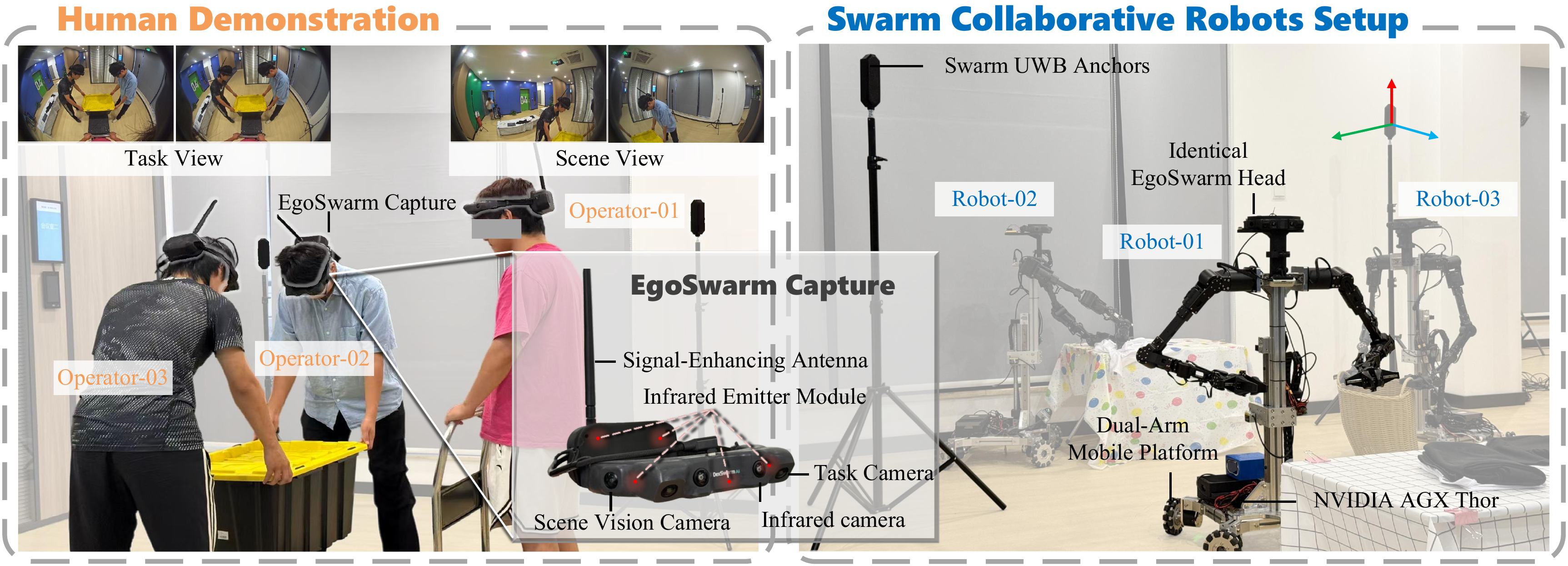}
\caption{\textbf{EgoSwarm Capture and shared human--robot sensing interface.}
Human participants wear SwarmEgo sensing heads to record synchronized multi-person egocentric demonstrations. Each sensing head integrates task and scene cameras, infrared cameras, an IMU, and UWB sensing. The robot platform reuses the same sensing interface and UWB infrastructure, reducing the perception gap between human demonstrations and decentralized robot deployment.}
\label{fig:2}

\end{figure}

\section{Related Work}\label{sec:related}

\subsection{Egocentric Experience and Human-to-Robot Transfer}\label{sec:egocentric}

Egocentric research has expanded from modeling individual daily activities to understanding skills across viewpoints and three-dimensional hand--object interactions. Ego4D, Ego-Exo4D, and HOT3D provide large-scale data on egocentric activities, cross-view behavior, and three-dimensional hand--object interactions, respectively \cite{ref8,ref9,ref10}. In human-to-robot learning, UMI and DexUMI explore robot-free collection of human manipulation demonstrations and cross-embodiment skill transfer \cite{ref15,ref16}. EgoMimic, HumanEgo, Ego2Robot, DexWild, and EgoZero further leverage human egocentric experience to learn robot representations and manipulation policies \cite{ref11,ref12,ref13,ref14,ref17}, while Open X-Embodiment advances robot learning across diverse tasks and embodiments \cite{ref6}. However, existing research still primarily focuses on individual skills and policies, with limited modeling of cross-agent behavioral relations in multi-person egocentric experience. EgoSwarm extends egocentric data collection to synchronized multi-person collaboration, providing a scalable source of interaction experience for learning coordinated multi-robot behavior.

\subsection{Swarm Intelligence and Decentralized Multi-Robot Learning}\label{sec:multiagent}

Swarm robotics emphasizes scalable collective behavior through local sensing and inter-agent interactions \cite{ref3,ref4,ref5}. Multi-agent learning commonly adopts centralized training with decentralized execution (CTDE). MADDPG, QMIX, TarMAC, and Multi-Agent Transformer study cooperation through policy optimization, value factorization, communication, and collective representations \cite{ref20,ref21,ref22,ref23}. Methods such as RoCo use large language models for multi-robot task decomposition and negotiation \cite{ref24,ref25}, while Omni-Swarm enables coordination through state estimation and local interactions \cite{ref26}. For real-world physical collaboration, CHORUS employs a shared VLA policy for decentralized collaboration across embodiments based on local observations, while SAI learns physically coupled two-robot behavior through asymmetric imitation \cite{ref29,ref30}. Related work further investigates closed-loop planning and execution \cite{ref28}. Nevertheless, existing methods still rely primarily on multi-agent data collected using robots or in simulation. The cost of data collection and the effort required to deploy and configure robots across different scenarios make it difficult to continuously expand collaborative experience, limiting the scalability and generalization of multi-robot policy learning.

\begin{figure}[tbp]
\centering
\includegraphics[width=\textwidth]{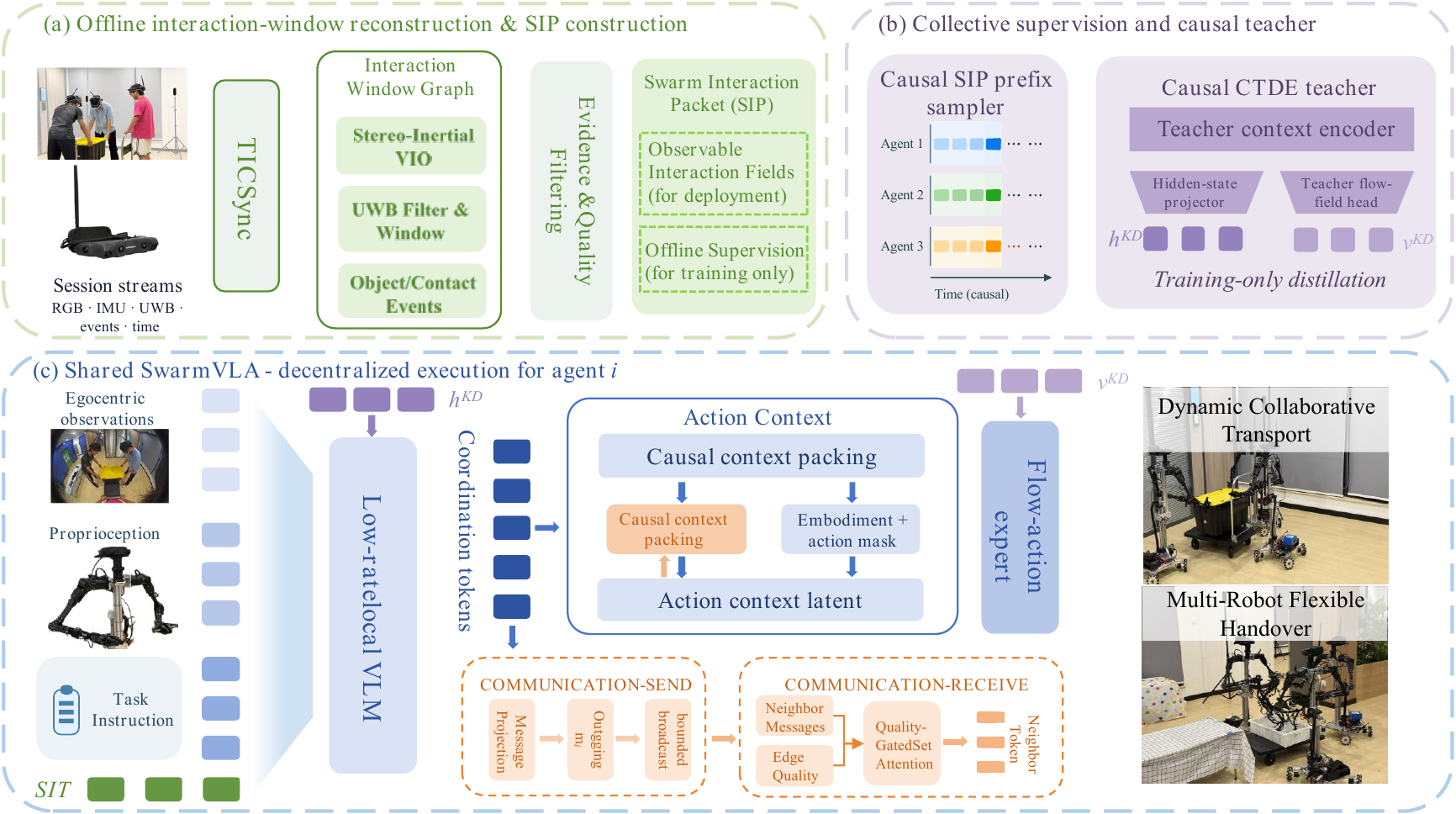}
\caption{\textbf{EgoSwarm pipeline for decentralized multi-robot collaboration.}
(a) Multimodal egocentric recordings are temporally aligned and reconstructed into quality-annotated Swarm Interaction Packets (SIPs).
(b) A causal CTDE teacher learns coordination representations and action flow fields from collective interaction supervision, which are distilled into the decentralized student policy.
(c) During deployment, each robot constructs local Swarm Interaction Tokens (SITs), aggregates quality-aware neighbor information, and generates future actions from the resulting Action Context.}
\label{fig:3}

\end{figure}

\section{EgoSwarm Method}\label{sec:method}

EgoSwarm learns decentralized multi-robot coordination from human collaborative experience. We introduce a robot-free multi-person egocentric data collection system (Sec.~\ref{sec:hardware}), reconstruct swarm interactions into structured SIP and SIT representations (Sec.~\ref{sec:processing}), and develop SwarmVLA to transfer this experience into decentralized policies using only local information (Sec.~\ref{sec:swarmvla}).

\subsection{Hardware Setup}\label{sec:hardware}

EgoSwarm uses a shared sensing interface for humans and robots to learn multi-robot collaboration policies from human swarm experience. This design enables collaborative demonstrations to be collected without robot hardware while preserving consistent observation and interaction information across data collection and deployment.

\textbf{EgoSwarm Capture.} EgoSwarm Capture is a robot-free multi-person egocentric data collection system that enables participants to directly manipulate real objects and collaborate naturally without deploying or teleoperating robots. As shown in Fig.~\ref{fig:2}, each participant wears a SwarmEgo sensing head and performs tasks within the coverage area of the Swarm UWB base stations. Each sensing head integrates six cameras, an inertial measurement unit (IMU), and an ultra-wideband (UWB) tag. Two infrared cameras operate together with distributed infrared markers to provide short-range relative observations between participants. The remaining four cameras synchronously record hands, partners, and shared objects while supporting visual-inertial motion estimation. UWB provides inter-agent ranging measurements and link-quality information to help identify candidate interaction windows. All sensor data and timestamps are logged independently on each device, providing raw observations for subsequent cross-device temporal alignment and reconstruction of local interaction relations.

\textbf{Multi-robot platform.} To reduce perceptual differences between human demonstrations and robot execution, we build a multi-robot platform that reuses the SwarmEgo sensing head. A shared mounting interface allows the robots to retain the camera, IMU, UWB, and time-synchronization configurations used during human data collection. Humans and robots therefore share a consistent sensing interface and interaction fields. The robots additionally record proprioceptive states and control commands to establish correspondences between interaction states and the robot action space. Robot adaptation data and action mapping account for human--robot kinematic differences, while relative geometry, shared-object information, contact relations, and message quality are represented in a common form. Each robot carries an NVIDIA AGX Thor and independently performs local perception, online Packetizer processing, SwarmVLA inference, and neighbor communication, enabling decentralized closed-loop execution without a central server.

\subsection{Swarm Interaction Data Processing}\label{sec:processing}

EgoSwarm converts synchronized multi-person egocentric recordings into structured swarm interaction supervision for SwarmVLA. As shown in Fig.~\ref{fig:3}(a), distributed RGB, infrared, IMU, and UWB observations are processed through temporal synchronization, interaction reconstruction, quality estimation, and action mapping. The resulting information is then organized into Swarm Interaction Packets (SIPs) with validity and quality annotations.

\textbf{Temporal Synchronization.} SwarmEgo devices operate on independent clocks, and fixed clock offsets and frequency drift can disrupt temporal correspondence between events observed by different participants. EgoSwarm adopts TICSync-style time mapping \cite{ref18} to map continuous recordings from each device onto a common session timeline, providing a shared temporal reference for associating observations across participants.

\textbf{Interaction Window Reconstruction.} Visual-inertial simultaneous localization and mapping (SLAM) first constructs a local map and estimates a motion trajectory for each participant \cite{ref19}. UWB-based proximity, interaction duration, and link quality are used to identify candidate interaction windows, which are then validated using overlapping views, shared objects, or contact evidence. Because the local maps have different origins and orientations, the system selects the SLAM map of one participant as a temporary reference frame within each interaction window and jointly refines the local-map transformations and residual time offsets:

\begin{equation}
\begin{aligned}
(\widehat{Y}^{k},\widehat{\delta}^{k})
 &=\underset{Y,\delta}{\operatorname{arg\,min}}
 \sum_{\ell\in\mathcal{F}_k}
 \rho_\ell\!\left(\lVert r_\ell(Y,\delta)\rVert_{\Sigma_\ell^{-1}}^{2}\right),\\
 &\qquad Y_r=I,\quad\delta_r=0.
\end{aligned}
\label{eq:reconstruction}
\end{equation}

Here, $Y_{i}$ transforms coordinates from the local map of participant $i$ into the reference frame, and $\delta {i}$ denotes the residual time offset. The map transformation and time offset of the reference participant are fixed to establish the spatial and temporal reference. $\mathcal{F}{k}$ contains the available visual associations, infrared-marker observations, UWB range measurements, shared-object constraints, and contact constraints. $\Sigma _{\ell }$ and $\rho _{\ell }$ characterize measurement uncertainty and robust treatment of residuals, respectively.

\textbf{Interaction Annotation and Quality Estimation.} EgoSwarm assigns validity, uncertainty, and evidence levels to reconstructed interaction relations based on the observations that support them. Fields with insufficient measurement constraints are downgraded or masked. Using the reconstructed interaction graph, the system identifies interaction events such as waiting, handovers, joint manipulation, failures, and recovery behaviors from relative motion, changes in object possession, and contact transitions. A frozen annotation model provides labels for roles, task stages, shared objects, and contact states, which are further verified through geometric consistency checks and manual spot checks.

\textbf{Cross-Embodiment Action Retargeting.} Human actions are derived from hand and shared-object motion. Hand keypoints define a virtual gripper pose and aperture, which are mapped to robot end-effector targets through embodiment-specific action mapping.

\textbf{SIP Construction and Causal Sampling.} Each valid interaction window is organized into an SIP. To strictly separate information available during policy execution from supervision used only during offline training, SIP fields are divided into three categories:

\textbf{1) Observable Interaction Fields.} These fields include relative geometry supported by currently available observations, inter-agent distances, shared objects, contact states, and the validity and quality of interaction relations. They correspond to the online sensing interface available during robot deployment and are further encoded into SITs as policy inputs.

\textbf{2) Offline Supervision Fields.} These fields include action targets, roles, task stages, interaction events, and quality information associated with the complete interaction window. They are used only for teacher learning, training-target construction, and sample selection, and are not provided as policy inputs during deployment.

\textbf{3) Causal Training Prefix.} Each training sample is bounded by a decision time. Policy inputs are constructed using only local history, neighbor messages, and relation quality available before that time. The complete interaction window is used only to generate supervision labels and assess sample quality, thereby preventing future information leakage. Segments without valid evidence of swarm interaction are retained as Human Individual Operation and used to adapt the model to individual manipulation.

SIPs thus convert distributed multi-person egocentric recordings into learning samples that are temporally and spatially aligned, explicitly encode interaction relations, and satisfy causal input constraints. They provide a common interface for subsequent SIT construction and decentralized policy learning with SwarmVLA.

\subsection{SwarmVLA: Decentralized Swarm Policy}\label{sec:swarmvla}

SwarmVLA learns a shared-parameter decentralized policy from SIP supervision. As shown in Fig.~\ref{fig:3}(c), each robot encodes local visual history, proprioception, language, and online SITs, then aggregates neighbor information into an Action Context for future action prediction. Training uses offline-reconstructed swarm information to learn collaborative relations, while deployment relies only on local observations and neighbor messages.

\textbf{Local Interaction Encoding:} SIT compresses the interaction fields in SIP that are available at decision time into a local representation that robots can use online. For robot $i$ and its neighbor $j$, we define:

\begin{equation}
c_{ij}^{t}=\phi_{\mathrm{SIT}}\!\left(r_{ij}^{t},s_{ij}^{t},u_{ij}^{t},q_{ij}^{t},\Delta t_{ij}^{t};v_{ij}^{t}\right).
\label{eq:sit}
\end{equation}

Here, $r_{ij}^{t}$ encodes local relative geometry and inter-agent distance; $s_{ij}^{t}$ represents shared-object information and contact states; $u_{ij}^{t}$ denotes the normalized uncertainty of the interaction relation; and $q_{ij}^{t}$ and $\Delta t_{ij}^{t}$ represent communication quality and message age, respectively. The field mask $v_{ij}^{t}$ indicates which components are available. All geometric relations are expressed in the coordinate frame of the receiving robot. Missing fields are explicitly masked, while their validity annotations are retained to distinguish unavailable information from true zero values.

At deployment, the local Packetizer constructs SITs online from current sensor observations and received neighbor messages. UWB provides inter-agent distances and link status; visual and infrared observations recover relative geometry, shared-object information, and contact relations; and local history preserves the temporal context of interactions. This process converts the swarm interaction information represented in offline SIPs into local state representations that remain available under the causal information constraints of deployment.

\textbf{Quality-Aware Neighbor Aggregation:} Geometric uncertainty, message delay, and link quality jointly determine the reliability of neighbor information. The vision-language module integrates local features, SITs, and interaction history into a Coordination Token and broadcasts a fixed-dimensional representation. The receiving robot then computes an aggregation weight based on relation quality and message age:

\begin{equation}
\begin{aligned}
\gamma_{ij}^{t}
 &=\exp\!\left(-\lambda_u u_{ij}^{t}-\lambda_t\Delta t_{ij}^{t}\right)\\
 &\quad\cdot\operatorname{sigmoid}\!\left[\lambda_q\left(q_{ij}^{t}-q_0\right)\right].
\end{aligned}
\label{eq:quality}
\end{equation}

The nonnegative coefficients $\lambda _{u},\lambda {t},\lambda {q}$ control the contributions of the corresponding quality factors, and $q{0}$ denotes the link-quality threshold. Quality-Gated Set Attention uses these weights to aggregate SITs and messages from valid neighbors into a Neighbor Token, denoted by $n{i}^{t}$. Invalid or stale messages are masked before attention normalization, and a learned null token is used when no valid neighbors are available. The Neighbor Token, local vision-language features, Coordination Token, proprioceptive state, and embodiment description are jointly combined to form the Action Context.

\textbf{Flow-Based Action Generation:} The Action Context is mapped through a conditioning adapter to the Flow-Action Expert, which generates an action chunk over the future prediction horizon. The model is initialized from pretrained weights \cite{ref2}. We jointly fine-tune the vision-language backbone, action head, and interaction modules, while training the action expert with a flow matching objective \cite{ref27}.

\begin{figure}[tbp]
\centering
\includegraphics[width=0.55\linewidth]{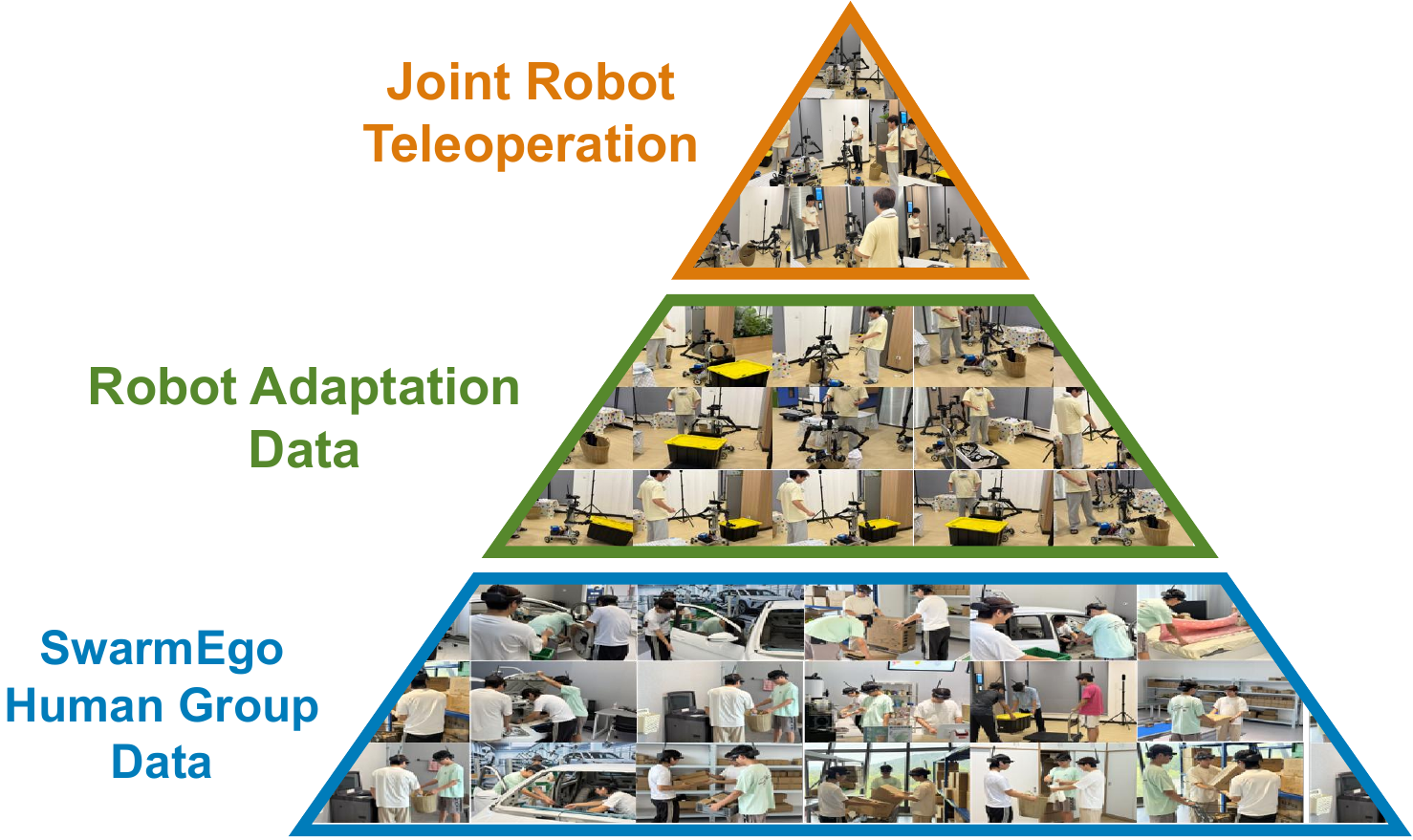}
\caption{\textbf{Composition of the SwarmVLA training data.}
The dataset consists of three complementary sources for decentralized multi-robot learning: human swarm data for cross-agent interaction experience, robot adaptation data for embodiment alignment, and joint robot teleoperation data for task-specific collaborative action supervision.}
\label{fig:4}

\end{figure}

\begin{figure}[tbp]
\centering
\includegraphics[width=\textwidth]{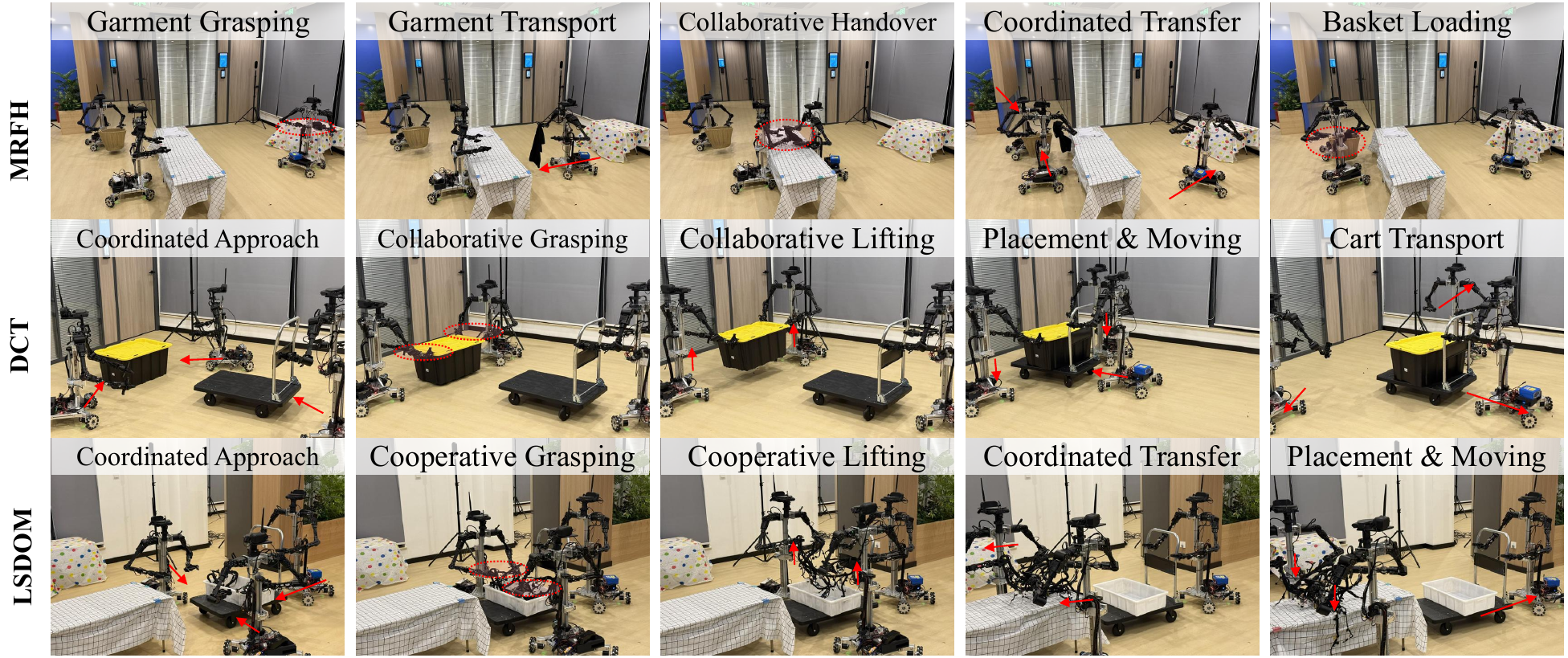}
\caption{\textbf{Real-world multi-robot collaborative task execution.}
The figure illustrates representative execution sequences of the three evaluation tasks: Multi-Robot Flexible Handover (MRFH, top), Dynamic Collaborative Transport (DCT, middle), and Large-Scale Deformable Object Manipulation (LSDOM, bottom). Frames progress from left to right; red arrows indicate robot motion directions, and dashed circles highlight key interaction regions.}
\label{fig:5}
\end{figure}

\textbf{Swarm-to-Local Distillation:} SwarmVLA distills offline-reconstructed swarm information into decentralized policies using only local information. Training combines human swarm recordings for collaborative behaviors, robot adaptation data for perception-action alignment, and task-specific joint teleoperation for additional action supervision. Dataset details are provided in the experimental setup.

Training follows three stages. Stage I adapts the model to individual manipulation using robot adaptation data and Human Individual Operation. Stage II trains a swarm teacher on human swarm interactions and robot demonstrations. Stage III freezes the teacher and distills its coordination representations and action flow fields into a shared student policy. Message delays, packet loss, and missing fields are simulated during training for communication robustness.

At deployment, the swarm teacher and offline SIP reconstructor are removed. Each robot runs the shared SwarmVLA policy with a local Packetizer and acts independently from local visual history, proprioception, task instructions, and limited neighbor messages, enabling decentralized closed-loop execution.

\section{Experiments}\label{sec:experiments}

We evaluate EgoSwarm through real-world multi-robot experiments, focusing on four questions:

\begin{itemize}

\item \textbf{Q1.} Does EgoSwarm improve decentralized collaboration performance?

\item \textbf{Q2.} Does human swarm experience provide additional gains under a fixed robot-data budget?

\item \textbf{Q3.} How do the interaction representation and quality-aware aggregation contribute, and how sensitive is performance to reduced human swarm data and degraded communication?

\item \textbf{Q4.} Do spatiotemporal interaction reconstruction and online inference provide the accuracy and real-time performance required for real-world deployment?

\end{itemize}

\subsection{Dataset and Experimental Setup}\label{sec:dataset}

As shown in Fig.~\ref{fig:4}, EgoSwarm Capture has collected approximately 50 hours of synchronized human collaboration data across 30 real-world scenario categories, covering diverse objects, participants, roles, and collaboration patterns. The dataset also captures natural interaction events, including waiting, interrupted handovers, role adjustments, and failure recovery, demonstrating the scalability of EgoSwarm Capture.

We evaluate EgoSwarm on three representative multi-robot tasks requiring continuous adaptation to partners, shared objects, and task progress. \textbf{1) Multi-Robot Flexible Handover (MRFH):} Three robots sort laundry through coordinated handovers and basket loading, testing conditional handovers and temporal coordination. \textbf{2) Dynamic Collaborative Transport (DCT):} Two robots load a box onto a cart before a third takes over transport, testing joint support and role transitions. \textbf{3) Large-Scale Deformable Object Manipulation (LSDOM):} Multiple robots jointly manipulate a wire harness, testing sustained coordination through a shared deformable object.

\subsection{Multi-Robot Deployment Experiments}\label{sec:deployment}

For every experimental setting, each method or condition is evaluated in 30 independent trials on each task, and the reported success rate is computed over these 30 trials.

We compare EgoSwarm with four baselines: the rule-based Local-Rule, Independent VLA without communication, Dec-VLA-Comm with standard aggregation of neighbor messages, and Centralized Joint VLA with global observations. The centralized method serves only as a reference with access to global information. All learning methods use the same data splits, robot-data budget, and hardware configuration. To match communication conditions, Dec-VLA-Comm uses the same message size and communication frequency as EgoSwarm, but does not include the structured SIT interaction representation or quality-aware gating.

\begin{table}[tbp]
\caption{Success rates (\%) across three swarm manipulation tasks.}
\label{tab:main}
\centering
\small
\setlength{\tabcolsep}{2pt}
\renewcommand{\arraystretch}{1.15}
\begin{tabular*}{\columnwidth}{@{\extracolsep{\fill}}lcccc@{}}
\hline
\noalign{\vskip 4pt}
Method & MRFH & DCT & LSDOM & Mean \\[3pt]
\hline
Local-Rule & 20.0 & 26.7 & 10.0 & 18.9 \\
Independent VLA & 43.3 & 46.7 & 36.7 & 42.2 \\
Dec-VLA-Comm & 63.3 & 56.7 & 30.0 & 50.0 \\
Centralized Joint VLA & 80.0 & 50.0 & 46.7 & 58.9 \\
\textbf{EgoSwarm} & \textbf{80.0} & \textbf{73.3} & \textbf{63.3} & \textbf{72.2} \\
\hline
\end{tabular*}

\end{table}

As shown in Table~\ref{tab:main}, EgoSwarm achieves success rates of 80.0\%, 73.3\%, and 63.3\% on MRFH, DCT, and LSDOM, respectively, yielding a mean success rate of 72.2\%. It exceeds Independent VLA and Dec-VLA-Comm by 30.0 and 22.2 percentage points, respectively, showing that decentralized collaboration benefits from both inter-agent communication and explicit interaction modeling. EgoSwarm also exceeds the Centralized Joint VLA reference by 13.3 percentage points in mean success rate despite using only local observations and limited neighbor messages. These results support the effectiveness of transferring interaction knowledge from human swarm demonstrations to decentralized robot policies. LSDOM remains the most challenging task because of its sustained multi-point coupling and deformable-object dynamics.

\subsection{Ablation and Robustness Tests}\label{sec:ablations}

We evaluate the contributions and robustness of EgoSwarm on the same three tasks as the main experiments. Specifically, we ablate human swarm experience, SIT, and quality-aware aggregation, and further test sensitivity to reduced human data and unreliable neighbor communication. The following variants are considered:

\begin{itemize}
\item \textbf{Robot-only:} Removes human swarm data and trains only on robot adaptation and joint teleoperation data.
\item \textbf{50\% human swarm data:} Uses half of the human swarm demonstrations while retaining the full robot-data budget and original data splits.
\item \textbf{w/o SIT:} Removes SIT features from the local vision-language model and neighbor aggregation while retaining visual observations and basic messages.
\item \textbf{w/o quality gate:} Replaces continuous weighting based on uncertainty, message age, and link quality with binary validity masks.
\item \textbf{30\% loss + 100 ms delay:} Evaluates the fully trained SwarmVLA policy under 0.3 independent message loss and an additional 100 ms one-way delay, while retaining original timestamps and validity thresholds.
\end{itemize}

Except for the factor specified by each variant, the comparisons retain the same model initialization, training budget, parameter sharing, causal inputs, and message format and transmission rate. All ablation and robustness conditions follow the same 30-trial-per-task evaluation protocol as the main experiments. The communication stress test uses the same full-model checkpoint and task evaluation protocol as the nominal setting.

\begin{table}[tbp]
\caption{Success rates (\%) for ablation and robustness tests. }
\label{tab:ablation}
\centering
\small
\setlength{\tabcolsep}{2pt}
\renewcommand{\arraystretch}{1.15}
\begin{tabular*}{\columnwidth}{@{\extracolsep{\fill}}lcccc@{}}
\hline
\noalign{\vskip 4pt}
Variant / condition & MRFH & DCT & LSDOM & Mean \\[3pt]
\hline
Robot-only & 33.3 & 20.0 & 13.3 & 22.2 \\
50\% human swarm data & 63.3 & 46.7 & 50.0 & 53.3 \\
w/o SIT & 46.7 & 46.7 & 23.3 & 38.9 \\
w/o quality gate & 56.7 & 36.7 & 43.3 & 45.6 \\
30\% loss + 100 ms delay & 76.7 & 70.0 & 53.3 & 66.7 \\
\textbf{Full SwarmVLA} & \textbf{80.0} & \textbf{73.3} & \textbf{63.3} & \textbf{72.2} \\
\hline
\end{tabular*}

\end{table}

\textbf{Human swarm experience.} Full SwarmVLA achieves a mean success rate of 72.2\%, compared with 22.2\% for Robot-only, corresponding to a 50.0-point improvement under the same robot-data budget. Using 50\% of the human swarm demonstrations achieves 53.3\%, while the full dataset provides a further 18.9-point gain. These results indicate that human swarm demonstrations provide substantial additional supervision for learning coordinated behavior, with further benefits from increasing the amount of swarm experience.

\textbf{Interaction representation and aggregation.} Removing SIT reduces the mean success rate from 72.2\% to 38.9\%, a decrease of 33.3 percentage points, while replacing quality-aware weighting with binary validity masks reduces it to 45.6\%, a decrease of 26.6 points. These results indicate that structured interaction representation provides the larger contribution in this ablation, while reliability-aware aggregation is also important for effective decentralized collaboration.

\textbf{Communication robustness.} Under 30\% packet loss and an additional 100 ms delay, SwarmVLA achieves success rates of 76.7\%, 70.0\%, and 53.3\% on MRFH, DCT, and LSDOM, respectively, corresponding to a mean of 66.7\%. This is 5.5 percentage points below nominal performance, showing that the policy retains most of its success rate under substantial communication degradation and supporting the effectiveness of quality-aware aggregation under missing and delayed neighbor messages.

\begin{figure}[tbp]
\centering
\includegraphics[width=0.82\linewidth]{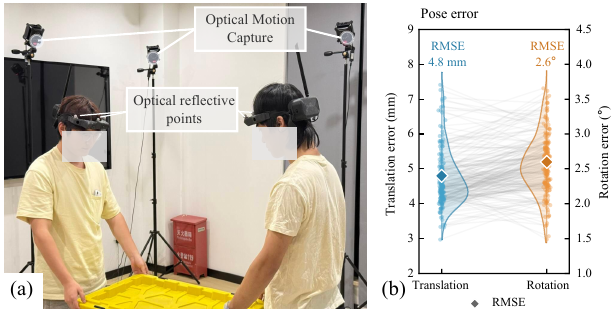}
\caption{\textbf{Evaluation of SIP relative-geometry reconstruction.}
(a) Optical motion capture provides reference poses for ground-truth evaluation.
(b) Translation and rotation error distributions over 30 independent multi-person interaction segments. SIP achieves a translation RMSE of 4.8\,mm and a rotation RMSE of 2.6$^\circ$.}
\label{fig:6}

\end{figure}

\subsection{SIP Accuracy and Runtime Performance}\label{sec:accuracy}

\textbf{Temporal synchronization accuracy.} We use 100 hardware-triggered events jointly observed by three head-worn sensing units and an independent recorder as reference events. These events are excluded from estimation of the time-mapping parameters. After synchronization, the mean absolute timestamp error relative to the reference timeline is 0.21 ms, the 95th-percentile (P95) error is 0.47 ms, and the maximum error is 1.08 ms. These results demonstrate stable submillisecond temporal alignment across devices, providing a reliable basis for associating visual observations, motion, and contact events across participants.

\textbf{Relative geometry accuracy.} We evaluate local relation reconstruction on 30 independent multi-person interaction segments using a 120 Hz optical motion-capture system as reference ground truth. The test segments cover representative interactions, including mutual approach, relative motion, handovers, and joint manipulation. Full SIP reconstruction achieves a translation root mean square error (RMSE) of 4.8 mm, a rotation RMSE of 2.6\ensuremath{^\circ}, and a valid-output ratio of 92.7\%. Compared with a baseline that directly fuses visual-inertial odometry (VIO) and UWB measurements, the translation and rotation errors are reduced by approximately 63.6\% and 67.9\%, respectively, indicating that multimodal constraints within interaction windows substantially improve the stability of local interaction geometry.

\textbf{Online runtime performance.} During collaborative execution with three robots, the local Packetizer requires an average of 2.2 ms per update, while SwarmVLA requires an average of 14.1 ms per policy inference. The system achieves an end-to-end P95 latency of 38.6 ms and consistently maintains closed-loop control at 30 Hz. The mean communication bandwidth of neighbor messages is approximately 11.8 KB/s per robot.

Overall, EgoSwarm reliably reconstructs local interaction relations with submillisecond temporal error and millimeter-scale spatial error while satisfying the computational and communication requirements for decentralized closed-loop execution at 30 Hz.

\section{Conclusion}\label{sec:conclusion}

We presented EgoSwarm, a framework that converts synchronized multi-person egocentric experience into local supervision for decentralized multi-robot collaboration. With EgoSwarm Capture, SIP/SIT representations, and SwarmVLA, it learns cross-agent relations from human swarm experience and executes using only local observations and limited neighbor messages. Across three real-world multi-robot task categories, EgoSwarm achieves a mean success rate of 72.2\%, supporting the contributions of human swarm experience, structured interaction representations, and quality-aware aggregation.

\textbf{Limitations and Future Work.} The current system mainly relies on visual, inertial, and UWB signals, reconstructs interactions only within sufficiently observed local windows, and is evaluated primarily with three robots. Future work will incorporate tactile and force sensing to model contact transitions, force transmission, and dynamic object states, and extend to larger teams, dynamic membership, and degraded communication for improved robustness and open-world generalization.


\clearpage

\begin{center}
    {\LARGE\bfseries Supplementary Materials\par}
\end{center}

\vspace{1.5em}

{
    \renewcommand{\contentsname}{\Large Contents}
    \tableofcontents
}

\clearpage
\appendix

\counterwithin{figure}{section}
\counterwithin{table}{section}
\renewcommand{\thefigure}{\thesection\arabic{figure}}
\renewcommand{\thetable}{\thesection\arabic{table}}

\section{Hardware and Data Collection Details}
\label{app:hardware}

\begin{figure}[h]
\centering
\includegraphics[width=0.95\linewidth]{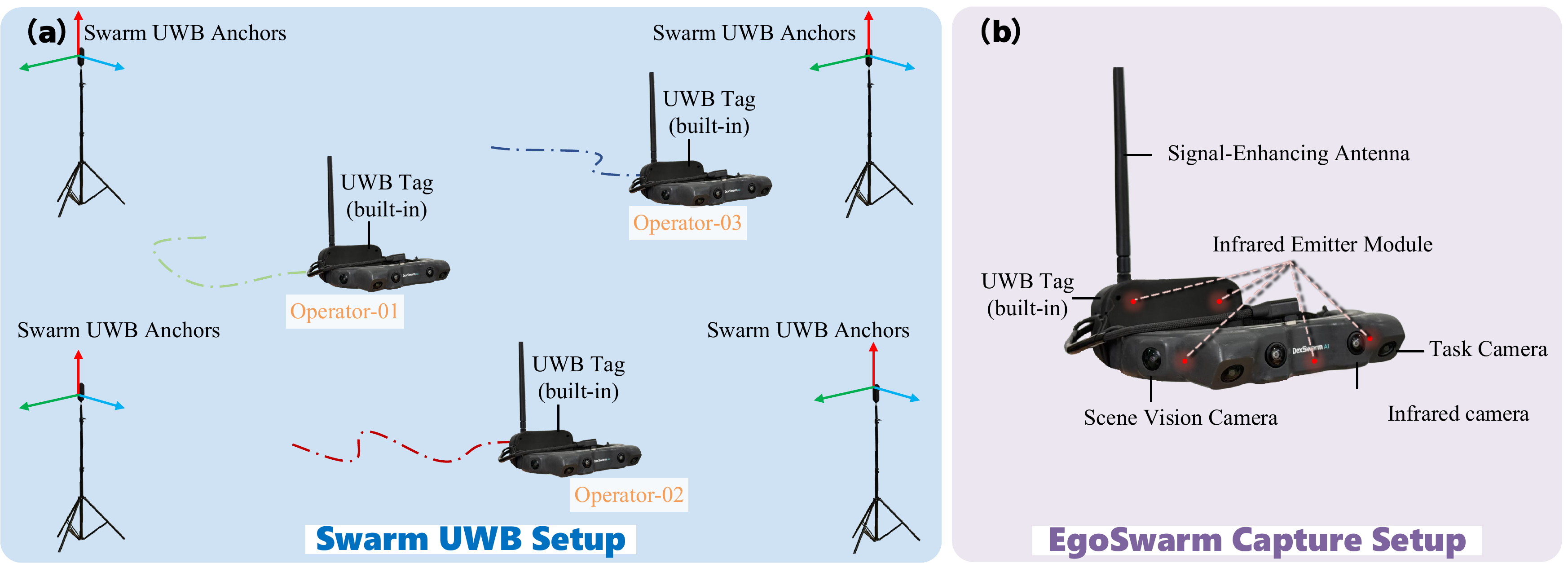}
\caption{\textbf{Multi-person data collection setup and SwarmEgo sensing hardware.}(a) The multi-person capture environment is equipped with distributed UWB anchors for inter-agent ranging, with each participant wearing an identical SwarmEgo sensing head. (b) The SwarmEgo device integrates RGB cameras, infrared cameras, an IMU, and UWB sensing to capture multimodal egocentric observations during natural human collaboration.}
\label{fig:6}

\end{figure}
\subsection{EgoSwarm Capture Hardware}
The SwarmEgo sensing head is a wearable multi-camera system designed to capture natural human collaboration from egocentric perspectives. As illustrated in Fig. A1, each device integrates four RGB cameras, two infrared cameras, an inertial measurement unit (IMU), and an ultra-wideband (UWB) tag. The RGB cameras capture hand movements, shared objects, and interactions with neighboring participants, while the infrared cameras operate with distributed infrared emitters to support short-range relative observations. The IMU provides inertial measurements for motion estimation, and the UWB module supplies inter-agent ranging and link-quality information.

All participants use an identical sensing configuration, providing a consistent multimodal observation interface across human demonstrations. The integrated design enables robot-free collection of multi-person interaction experience without restricting natural collaborative movements.

\subsection{Multi-Person Data Collection and Synchronization}
As shown in Fig. A1, the multi-person data collection system consists of multiple participants equipped with SwarmEgo sensing heads and a set of spatially distributed UWB anchors. Participants perform collaborative manipulation tasks naturally, including coordinated transport, object handovers, and shared-object manipulation, without requiring robot hardware or teleoperation interfaces. The UWB infrastructure provides ranging measurements that support the identification of spatial proximity and potential interaction between participants.

Each SwarmEgo device independently records RGB, infrared, IMU, and UWB observations with associated timestamps. Since the devices operate on independent clocks, their recordings may exhibit temporal offsets and clock drift. To establish a consistent temporal reference, we adopt TICSync-style time mapping to align individual sensor streams onto a shared session timeline. The synchronized recordings are subsequently used for interaction-window reconstruction and Swarm Interaction Packet (SIP) construction, as described in Sec. 3.2.

\subsection{Multi-Robot Platform and Deployment}

\begin{figure}[h]
\centering
\includegraphics[width=0.6\linewidth]{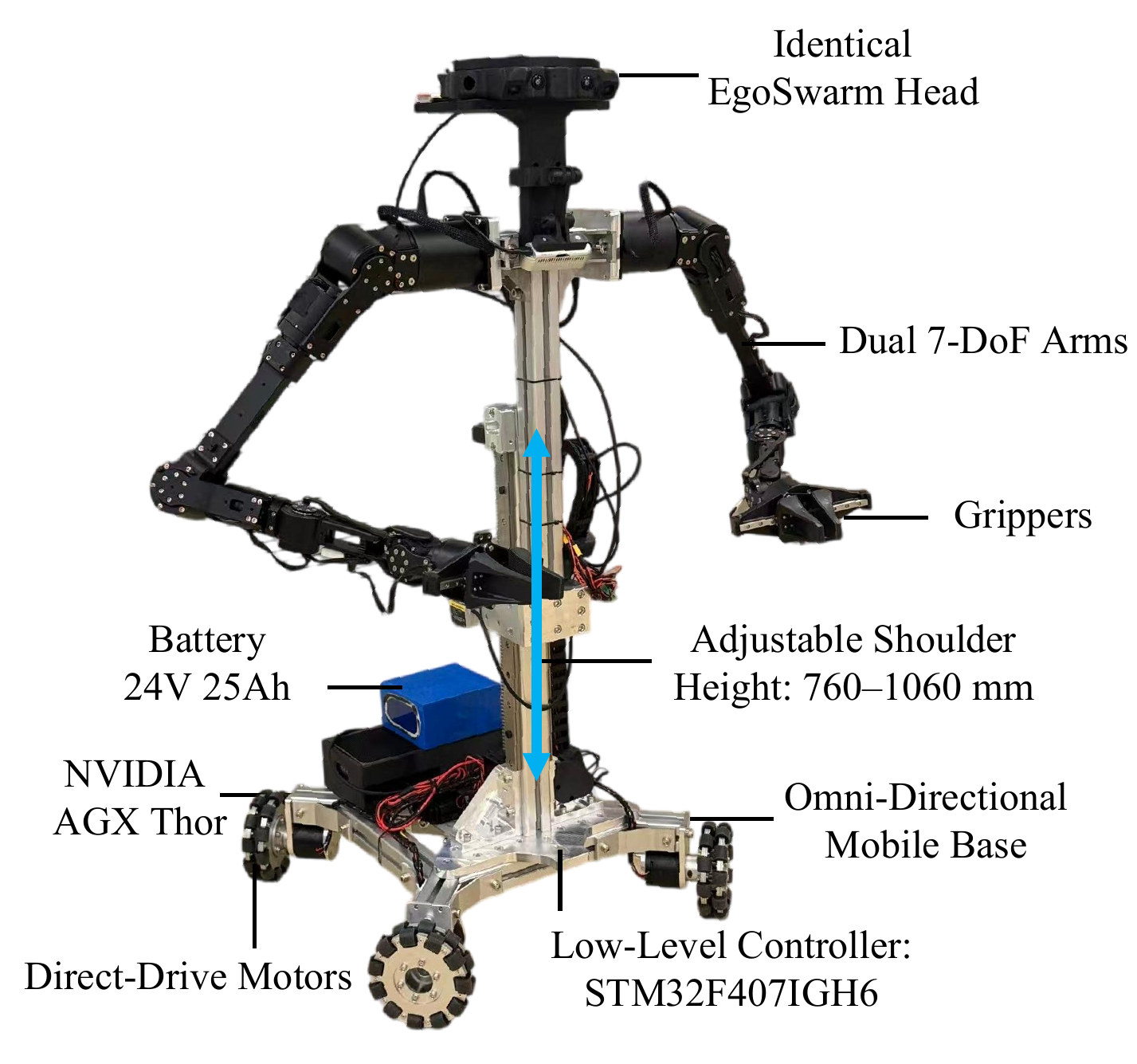}
\caption{\textbf{Multi-robot hardware platform.}Each robot integrates two 7-DoF manipulators, grippers,
an omnidirectional mobile base, and a vertical lifting
mechanism with a 300 mm stroke. An identical EgoSwarm
Head provides a shared sensing interface with human
demonstrations. An NVIDIA AGX Thor and an
STM32F407IGH6-based controller support onboard
computation and low-level actuation, respectively,
enabling decentralized robot execution.}
\label{fig:6}

\end{figure}

As illustrated in Fig. A2, we develop a mobile dual-arm robotic platform for real-world multi-robot collaboration. Each robot is equipped with two 7-DoF manipulators, grippers, and an omnidirectional mobile base driven by direct-drive motors. The platform incorporates a vertical lifting mechanism with a 300 mm stroke and an adjustable shoulder height of 760–790 mm, enabling manipulation at different working heights. A 24 V, 25 Ah battery provides onboard power. The computing and control system consists of an NVIDIA AGX Thor for high-level computation and an STM32F407IGH6-based controller for low-level actuation.

Each robot is equipped with an identical EgoSwarm Head, maintaining a consistent sensing interface between human swarm demonstrations and robot deployment. During execution, each robot independently performs local perception, online Packetizer processing, and SwarmVLA inference on its onboard computing system. Limited neighbor communication supports coordination without centralized policy execution, enabling decentralized closed-loop collaboration.

\section{Training Details}
\label{app:training-details}

\setcounter{table}{0}
\renewcommand{\thetable}{\thesection\arabic{table}}

\subsection{Training Configuration}
\label{app:training-configuration}

\paragraph{Optimization.}
We initialize SwarmVLA from the pretrained $\pi_{0.5}$ weights and train it on a compute node equipped with four NVIDIA A100 GPUs, each with 80\,GB of memory, a 64-core CPU, and 256\,GB of system memory. We use AdamW with $(\beta_1,\beta_2)=(0.9,0.95)$, weight decay of $0.01$, and $\epsilon=10^{-8}$. Each training stage begins with 1,000 steps of linear warmup to a peak learning rate of $2.5\times10^{-5}$, followed by cosine decay to $2.5\times10^{-6}$. A per-GPU micro-batch size of 2 and eight gradient accumulation steps yield an effective global batch size of 64. Training uses bfloat16 mixed precision, gradient checkpointing, and parameter sharding, with the gradient norm clipped at 1.0.

Each of the three stages runs for 30,000 steps. The vision-language backbone, action head, and interaction modules are jointly optimized; the swarm teacher is frozen during Stage III. The student is trained with action flow matching, coordination representation distillation, and action flow-field distillation. The overall distillation weight is $\lambda_{\mathrm{KD}}=1.0$, and the relative weight of representation distillation is $\lambda_h=0.1$. At inference time, action chunks are generated using 10 Euler integration steps. Table~\ref{tab:app-training-config} summarizes the optimization and inference settings.

\begingroup
\small
\setlength{\tabcolsep}{6pt}
\renewcommand{\arraystretch}{1.08}
\begin{longtable}{@{}>{\raggedright\arraybackslash}p{0.55\linewidth}>{\raggedright\arraybackslash}p{\dimexpr0.45\linewidth-12pt\relax}@{}}
\caption{Training hyperparameters and optimization settings for SwarmVLA. The learning-rate schedule is applied separately in each stage.}
\label{tab:app-training-config}\\
\toprule
\textbf{Parameter} & \textbf{Value} \\
\midrule
\endfirsthead
\multicolumn{2}{l}{\tablename~\thetable\ (continued)}\\
\toprule
\textbf{Parameter} & \textbf{Value} \\
\midrule
\endhead
\midrule
\multicolumn{2}{r}{Continued on the next page}\\
\endfoot
\bottomrule
\endlastfoot
Random seed & 42 \\
GPU configuration & $4\times$ NVIDIA A100 (80\,GB each) \\
Numerical precision & bfloat16 mixed precision \\
Per-GPU micro-batch size & 2 \\
Gradient accumulation steps & 8 \\
Effective global batch size & 64 \\
Optimizer & AdamW \\
AdamW $(\beta_1,\beta_2)$ & $(0.9,0.95)$ \\
Optimizer $\epsilon$ & $1\times10^{-8}$ \\
Weight decay & 0.01 \\
Peak learning rate & $2.5\times10^{-5}$ \\
Minimum learning rate & $2.5\times10^{-6}$ \\
Learning-rate schedule & Linear warmup followed by cosine decay \\
Warmup steps per stage & 1,000 \\
Stage I training steps & 30,000 \\
Stage II training steps & 30,000 \\
Stage III training steps & 30,000 \\
Gradient-norm clipping threshold & 1.0 \\
Overall distillation weight $\lambda_{\mathrm{KD}}$ & 1.0 \\
Relative representation-distillation weight $\lambda_h$ & 0.1 \\
Inference integrator & Euler \\
Inference integration steps & 10 \\
\end{longtable}
\endgroup

\paragraph{Inputs and model architecture.}
All input images are resized to $224\times224$ pixels. For each camera view, the model uses the current frame and the immediately preceding frame. The vision-language backbone and action expert retain hidden dimensions of 2,048 and 1,024, respectively. The SIT encoder maps structured interaction fields and their validity masks to a 256-dimensional representation. The Coordination Token is further compressed into a 128-dimensional message for communication with neighboring agents. At the receiver, each message is concatenated with its corresponding SIT representation, projected, and processed by a two-layer Quality-Gated Set Attention module to produce a 256-dimensional Neighbor Token. This token is fused with local vision-language features, the Coordination Token, proprioceptive state, and the embodiment description to form the conditioning context for the action expert.

The action head predicts chunks of 50 time steps through a unified 32-dimensional output interface. An embodiment-specific mask selects the valid action channels. The interaction modules use GELU activations, with a dropout rate of 0.1 in both the attention and feed-forward layers.

\paragraph{Baseline configurations.}
All learning-based baselines use the same data splits, robot-data budget, and hardware configuration. Dec-VLA-Comm matches SwarmVLA in message dimensionality and transmission frequency, but aggregates neighbor information using standard attention. Independent VLA uses local observations and task instructions, whereas Centralized Joint VLA receives global observations from multiple robots and serves as a centralized reference.

\subsection{Data Composition and Training Stages}
\label{app:data-and-stages}

\paragraph{Data composition.}
Training draws on three complementary data sources: human swarm demonstrations, robot adaptation data, and joint robot teleoperation data. Human swarm demonstrations provide cross-agent interaction experience, including waiting, handovers, joint manipulation, role transitions, and recovery from failures. Robot adaptation data support alignment between observations and embodiment-specific actions, while joint robot teleoperation data provide task-specific supervision for coordinated robot actions. Human recording segments without valid evidence of swarm interaction are retained as \emph{Human Individual Operation} data for individual manipulation adaptation.

Human swarm recordings are reconstructed into Swarm Interaction Packets (SIPs) with validity and quality annotations. For each training sample, policy inputs are restricted to the local observation history, neighbor messages, and interaction information available at the decision time. Information from the complete interaction window is used only to construct offline supervision and assess sample quality. This separation preserves the causal information constraints of decentralized deployment.

\paragraph{Stage I: individual adaptation.}
We first adapt the pretrained model to individual manipulation using robot adaptation data and Human Individual Operation data. This stage establishes perception--action alignment before learning cross-agent coordination.

\paragraph{Stage II: swarm teacher learning.}
We train a swarm teacher using human swarm interactions and robot demonstrations. The teacher uses offline-reconstructed swarm information to learn coordination representations and action flow fields under the centralized-training, decentralized-execution framework.

\paragraph{Stage III: shared student distillation.}
We freeze the teacher and distill its coordination representations and action flow fields into a shared-parameter student policy. The student combines action flow matching with the two distillation objectives described in Section~\ref{app:training-configuration}, while conditioning on locally available information. During training, we simulate message delays, packet loss, and missing interaction fields to improve robustness to unreliable communication and incomplete observations. At deployment, the swarm teacher and offline SIP reconstructor are removed. Each robot executes the shared student policy with a local Packetizer, using local observations, local history, and limited neighbor messages.

\subsection{Task Instructions and Prompts}
\label{app:task-prompts}

Task instructions consist of a shared task-level prompt and a role-specific prompt for each robot. The task-level prompt specifies the collective objective and completion conditions, while the role-specific prompt defines the responsibilities of an individual robot. Each robot combines these instructions with local observations and neighbor information to determine its actions and their timing. Tables~\ref{tab:app-task-prompts} and~\ref{tab:app-role-prompts} provide examples for Multi-Robot Flexible Handover (MRFH), Dynamic Collaborative Transport (DCT), and Large-Scale Deformable Object Manipulation (LSDOM).

\begingroup
\small
\setlength{\tabcolsep}{6pt}
\renewcommand{\arraystretch}{1.12}
\begin{longtable}{@{}>{\raggedright\arraybackslash}p{0.12\linewidth}>{\raggedright\arraybackslash}p{\dimexpr0.88\linewidth-12pt\relax}@{}}
\caption{Example task-level prompts specifying shared objectives and completion conditions.}
\label{tab:app-task-prompts}\\
\toprule
\textbf{Task} & \textbf{Task-level prompt} \\
\midrule
\endfirsthead
\multicolumn{2}{l}{\tablename~\thetable\ (continued)}\\
\toprule
\textbf{Task} & \textbf{Task-level prompt} \\
\midrule
\endhead
\midrule
\multicolumn{2}{r}{Continued on the next page}\\
\endfoot
\bottomrule
\endlastfoot
MRFH & Work together to sort, hand over, and place garments into baskets. Coordinate each handover with the receiving partner's readiness, and place all target garments into their corresponding baskets. \\
\addlinespace[4pt]
DCT & Work together to load the box onto the cart and transport it to the designated area. Two robots jointly load the box, and a third robot takes over transport once loading is complete. \\
\addlinespace[4pt]
LSDOM & Grasp and support the wire harness at different locations. Coordinate the motion of the grasp points and adjust your actions to its deformation to move and place the harness in the designated target configuration. \\
\end{longtable}
\endgroup

\begingroup
\small
\setlength{\tabcolsep}{5pt}
\renewcommand{\arraystretch}{1.12}
\begin{longtable}{@{}>{\raggedright\arraybackslash}p{0.11\linewidth}>{\raggedright\arraybackslash}p{0.22\linewidth}>{\raggedright\arraybackslash}p{\dimexpr0.67\linewidth-20pt\relax}@{}}
\caption{Example role-specific prompts for the three collaborative tasks. A, B, and C denote the participating robots.}
\label{tab:app-role-prompts}\\
\toprule
\textbf{Task} & \textbf{Role} & \textbf{Role-specific prompt} \\
\midrule
\endfirsthead
\multicolumn{3}{l}{\tablename~\thetable\ (continued)}\\
\toprule
\textbf{Task} & \textbf{Role} & \textbf{Role-specific prompt} \\
\midrule
\endhead
\midrule
\multicolumn{3}{r}{Continued on the next page}\\
\endfoot
\bottomrule
\endlastfoot
MRFH & A: Pick and hand over & Pick up a garment from the initial area and hand it to B. Release it only after B has established a stable grasp, then prepare the next garment. \\
\addlinespace[4pt]
  & B: Receive and relay & Receive the garment from A. After A releases it, pass it to C. Release it only after C has established a stable grasp. \\
\addlinespace[4pt]
  & C: Receive and place & Receive the garment from B. After B releases it, place it in the corresponding basket, then prepare to receive the next garment. \\
\midrule
DCT & A: Load from one side & Grasp one side of the box and coordinate with B to lift, move, and place it. Release your grasp only after the box rests stably on the cart. \\
\addlinespace[4pt]
  & B: Load from the other side & Grasp the opposite side of the box. Adjust your motion according to A's actions and the box pose, and release your grasp after jointly completing the loading operation. \\
\addlinespace[4pt]
  & C: Take over transport & Keep the cart stable during loading. Once the box is stably placed and A and B have cleared the transport path, move the cart to the target area. \\
\midrule
LSDOM & A: Manipulate one end & Grasp one end of the wire harness. Adjust your motion according to your partners' actions and the harness deformation to coordinate its transfer and the placement of your end. \\
\addlinespace[4pt]
  & B: Support the middle & Support the middle of the wire harness. Coordinate with both ends to adjust its position and shape, keep the harness smooth, and place the middle section. \\
\addlinespace[4pt]
  & C: Manipulate the other end & Grasp the opposite end of the wire harness. Coordinate your motion with A and B, maintain your grasp, and help place the harness in the target configuration. \\
\end{longtable}
\endgroup

\section{Baseline Failure Analysis}
\label{app:failure_analysis}

To better understand the challenges of multi-robot collaboration, we analyze representative failure modes associated with temporal coordination, joint support, and shared-object deformation. We further discuss how these failures relate to the information available to different baseline policies. Successful collaboration requires robots to account for partner readiness, object possession, and task progress when deciding when to wait, release, or take over an action.

\subsection{Temporal Misalignment and Failed Handovers}
\label{app:handover_failures}

\textit{Multi-Robot Flexible Handover.}
Garment handover requires the transferring and receiving robots to coordinate a transition in object possession. After bringing a garment into the handover region, the transferring robot must maintain support until the receiving robot establishes a stable grasp. The receiving robot, in turn, must approach and grasp the garment while it remains supported. The temporal relationship between these actions is critical to a successful handover.

Premature release can cause the garment to lose support before the transfer of possession is complete. If the transferring robot treats arrival at the handover position as task-stage completion, the garment may shift or deform before the receiving robot has secured it. Subsequent gripper closure may then miss the intended grasp location, resulting in a dropped garment or an interrupted handover. Conversely, the interaction may stall when both robots wait for the partner to initiate the next action or repeatedly adjust their local grasp configurations. Garment deformation, arm occlusion, and viewpoint differences further complicate the estimation of partner readiness.

Independent VLA relies on local visual observations and history to infer partner states, which may be insufficient when the interaction region is occluded. Dec-VLA-Comm incorporates neighbor messages through standard aggregation but does not explicitly structure object-possession relations or information reliability. Local-Rule advances actions according to predefined conditions, making its effectiveness dependent on how well the rules cover approach, contact, and waiting states. Across these baselines, the central challenge is to relate partner readiness to the completion conditions of each robot's own actions.

On MRFH, EgoSwarm achieves an 80.0\% success rate, compared with 43.3\% for Independent VLA and 63.3\% for Dec-VLA-Comm. EgoSwarm represents partner relations and contact states through the Swarm Interaction Token (SIT), incorporates quality-aware neighbor aggregation, and transfers cross-agent action dependencies through swarm-to-local distillation. These components provide a structured basis for coordinating reception, waiting, and release during collaborative handovers.

\subsection{Coordination Failures in Shared-Object Manipulation}
\label{app:shared_object_failures}

\textit{Dynamic Collaborative Transport.}
Box loading requires two robots to maintain joint support during grasping, lifting, and transfer, followed by a handover of transportation responsibility to a third robot. If one robot initiates lifting or translation while the other is still establishing its grasp, the resulting support imbalance may cause the box to tilt, slip, or require repositioning. Similarly, releasing support before the box is stably placed on the cart can interrupt the loading process.

Inconsistent task-stage transitions introduce an additional source of failure. Premature cart movement by the transporting robot may shift the placement target before loading is complete, whereas prolonged waiting after successful loading can delay subsequent transport. DCT therefore requires both continuous coordination during joint manipulation and appropriate transitions between loading and transportation roles. EgoSwarm achieves a 73.3\% success rate on this task, compared with 46.7\% for Independent VLA and 56.7\% for Dec-VLA-Comm. Explicit representations of shared-object interactions and task progress provide information relevant to these coordination requirements.

\textit{Large-Scale Deformable Object Manipulation.}
The wire harness physically couples multiple grasping points throughout manipulation. Motion at one location changes the configuration and available manipulation space at other locations, requiring each robot to adapt its actions to both object deformation and partner motion. If one robot moves ahead while the remaining grasping points fail to follow, the harness may become excessively stretched, potentially causing local slippage or loss of grasp. Incompatible motion directions can introduce crossings or entanglement, while repeated local corrections may prevent the team from making progress toward the target configuration.

Such tasks require sustained awareness of multi-point relations throughout execution. Independent VLA primarily generates actions from local observations, whereas standard neighbor-message aggregation must implicitly recover these relations from exchanged representations. SwarmVLA instead organizes relative geometry, shared-object information, contact states, and validity annotations into SITs. Quality-aware aggregation further adjusts the contribution of neighbor information according to its reliability and message age, allowing action generation to incorporate interaction context over time.

On LSDOM, EgoSwarm achieves a 63.3\% success rate. Removing SIT reduces this rate to 23.3\%, while removing quality-aware aggregation reduces it to 43.3\%. These ablations support the contributions of structured interaction representations and reliability-aware neighbor aggregation to successful deformable-object manipulation. Table~\ref{tab:failure_modes} summarizes the failure modes discussed above and the corresponding observation challenges.

\begin{table}[t]
\centering
\caption{\textbf{Failure modes and observation challenges in multi-robot collaboration.}
Representative coordination failures across the three evaluation tasks and the interaction information required to identify them.}
\label{tab:failure_modes}
\small
\renewcommand{\arraystretch}{1.15}
\setlength{\tabcolsep}{4pt}
\begin{tabularx}{\textwidth}{
    l
    p{2.6cm}
    X
    p{3.2cm}
}
\toprule
\textbf{Task} &
\textbf{Failure Mode} &
\textbf{Behavioral Manifestation} &
\textbf{Observation Challenge} \\
\midrule
MRFH &
Premature release &
The transferring robot releases the garment before the receiver establishes a stable grasp. &
Grasp, release, and object-possession relations. \\

  &
Prolonged waiting &
Robots remain in the handover region without advancing to the next task stage. &
Partner readiness and temporal coordination. \\

\midrule
DCT &
Interrupted joint support &
Asymmetric lifting or premature release causes box tilting or slippage. &
Bilateral end-effector motion and object pose. \\

  &
Mistimed role transition &
Cart movement begins prematurely or remains stalled after loading. &
Temporal relation between loading completion and transport initiation. \\

\midrule
LSDOM &
Grasp loss or local slippage &
Changing grasp-point configurations cause local slippage during harness transport. &
Contact regions and local deformation. \\

  &
Stalled coordinated manipulation &
Harness crossings, repeated corrections, or interrupted overall transfer. &
Multi-point motion and global object configuration. \\

\bottomrule
\end{tabularx}
\end{table}

\section{Extended Task Descriptions}
In this section, we provide additional details on the three real-world multi-robot collaborative tasks introduced in Sec.~4.1 and illustrated in Fig.~5. We describe the task configurations, execution stages, and evaluation criteria, with particular emphasis on the temporal coordination, role transitions, and shared-object interactions required for successful decentralized collaboration.

\subsection{Multi-Robot Flexible Handover (MRFH)}
\label{app:mrfh}

\textit{Initial Configuration.}
Three robots are deployed in a household laundry-sorting environment, where garments must be transferred between robots and placed into designated baskets. The robots begin from different positions around the workspace, requiring them to coordinate garment retrieval, transportation, handovers, and final placement. Unlike individual pick-and-place tasks, successful execution depends on the readiness and actions of neighboring robots, particularly during object transfer.

\textit{Step 1: Garment Grasping.}
The first robot approaches a garment and grasps it using its gripper. This stage requires the robot to establish a stable grasp on the flexible object before initiating transportation. \textit{Evaluation Criteria:} The garment must be securely grasped and retained during the subsequent movement.

\textit{Step 2: Garment Transport.}
After grasping, the robot transports the garment toward the collaborative handover region. The robot must account for the relative position and readiness of the receiving robot rather than following an independently timed trajectory. \textit{Evaluation Criteria:} The garment must reach the handover region while remaining securely held.

\textit{Step 3: Collaborative Handover.}
The receiving robot approaches the garment and establishes a stable grasp before the transferring robot releases it. This stage requires temporal coordination between the two robots, as premature release or delayed reception may result in a failed handover. \textit{Evaluation Criteria:} The garment must be successfully transferred without being dropped or losing control during the exchange.

\textit{Step 4: Coordinated Transfer.}
Following the handover, the receiving robot transports the garment toward its designated destination while the remaining robots coordinate their movements within the shared workspace. This stage evaluates whether the robots can maintain appropriate spatial and temporal coordination across successive actions. \textit{Evaluation Criteria:} The garment must be transferred toward the designated basket without unintended interference or loss of grasp.

\textit{Step 5: Basket Loading.}
Finally, the robot positions the garment above the designated basket and releases it to complete the sorting operation. \textit{Evaluation Criteria:} The task is considered successful when the required garment transfers are completed and the garment is placed into the designated basket without an unsuccessful handover or loss of the object.

\subsection{Dynamic Collaborative Transport (DCT)}
\label{app:dct}

\textit{Initial Configuration.}
Three robots collaborate to load a box onto a mobile cart and subsequently transport it. Two robots are initially responsible for manipulating the box, while the third robot takes over the transportation stage. The task involves both simultaneous physical cooperation and sequential role transitions, requiring the robots to coordinate their actions according to the progress of the shared manipulation.

\textit{Step 1: Coordinated Approach.}
Two robots approach the box from their respective positions and establish suitable configurations for collaborative manipulation. Their movements must be coordinated to avoid interference and prepare for joint grasping. \textit{Evaluation Criteria:} Both robots must reach appropriate manipulation positions without disrupting the box or obstructing one another.

\textit{Step 2: Collaborative Grasping.}
The two robots establish grasps on the box, preparing to support its weight during lifting. Since both robots interact with the same rigid object, the grasping actions must provide compatible support before the lifting motion begins. \textit{Evaluation Criteria:} Both robots must establish stable grasps that allow the box to be lifted collaboratively.

\textit{Step 3: Collaborative Lifting.}
The robots lift the box together while maintaining coordinated motion. Differences in lifting timing or motion direction can cause the box to tilt or become unstable, making this stage dependent on the actions of both robots. \textit{Evaluation Criteria:} The box must be lifted and maintained in a stable configuration without being dropped.

\textit{Step 4: Placement and Moving.}
After lifting, the robots coordinate the movement and placement of the box onto the cart. Both robots must adjust their actions as the shared object approaches its destination and release it only after stable placement. \textit{Evaluation Criteria:} The box must be successfully placed on the cart without falling or becoming unstable during the transfer.

\textit{Step 5: Cart Transport.}
Once the box has been loaded, the third robot takes over the transportation stage and moves the cart toward the designated destination. This transition requires the task to progress from simultaneous box manipulation to a different collaborative role configuration. \textit{Evaluation Criteria:} The task is considered successful when the box is securely loaded and the cart completes the required transportation without losing the load.

\subsection{Large-Scale Deformable Object Manipulation (LSDOM)}
\label{app:lsdom}

\textit{Initial Configuration.}
Multiple robots collaboratively manipulate a large flexible wire harness toward a target configuration. Unlike rigid-object transport, the wire harness can deform continuously during manipulation, and the motion of one robot influences the object configuration experienced by the others. This task therefore requires sustained coordination throughout the manipulation process rather than synchronization at isolated handover events.

\textit{Step 1: Coordinated Approach.}
The robots approach the wire harness from different positions and prepare to manipulate the shared deformable object. Their initial configurations must allow coordinated access to the harness while avoiding interference between manipulators. \textit{Evaluation Criteria:} The robots must reach suitable grasping positions without disturbing the object in a way that prevents subsequent manipulation.

\textit{Step 2: Cooperative Grasping.}
The robots establish grasps at different locations along the wire harness. Because the object is deformable, grasping at one location may change the positions of other segments and affect the remaining robots. \textit{Evaluation Criteria:} The robots must establish and maintain stable grasps that support subsequent coordinated manipulation.

\textit{Step 3: Cooperative Lifting.}
The robots lift the wire harness together while adapting to its changing shape. Uncoordinated lifting can introduce excessive deformation or uneven motion, making synchronization essential for maintaining control of the shared object. \textit{Evaluation Criteria:} The wire harness must be lifted while maintaining stable grasping and coordinated support among the robots.

\textit{Step 4: Coordinated Transfer.}
The robots transport and adjust the wire harness toward the target region. During this stage, each robot must account for the motion of neighboring robots and the changing configuration of the object, since independently generated movements may produce conflicting displacements. \textit{Evaluation Criteria:} The wire harness must be transferred toward the target configuration without losing grasp or causing a task-disrupting deformation.

\textit{Step 5: Placement and Moving.}
Finally, the robots coordinate the placement of the wire harness into the target configuration. The robots must adjust their relative movements and release the object in a coordinated manner to avoid disturbing previously positioned segments. \textit{Evaluation Criteria:} The task is considered successful when the wire harness reaches the required target configuration and the robots complete the placement without losing control of the shared object.

\end{document}